\documentclass{article}
\usepackage{ijcai20}

\usepackage{times}
\usepackage{soul}
\usepackage{url}
\usepackage[hidelinks]{hyperref}
\usepackage[utf8]{inputenc}
\usepackage[small]{caption}
\usepackage{graphicx}
\usepackage{amsmath}
\usepackage{amsthm}
\usepackage{amssymb}
\usepackage{mathrsfs}
\usepackage{booktabs}
\usepackage{algorithm}
\usepackage{algorithmic}
\usepackage{multirow}

\usepackage{color}

\usepackage{pifont}
\usepackage{array}
\title{Discovering Protagonist of Sentiment with Aspect Reconstructed Capsule Network}

\author{
Chi Xu$^{1,}$\footnotemark[6]
\and
Hao Feng $^{2,}$\footnotemark[6]
\and
Guoxin Yu$^3$\and
Min Yang$^4$ \and
Xiting Wang$^5$ \and
Xiang Ao$^{1,}$\footnotemark[7]
\affiliations
$^1$Institute of Computing Technology, Chinese Academy of Sciences\\
$^2$University of Electronic Science and Technology of China\\
$^3$Shandong Normal University\\
$^4$SIAT, Chinese Academy of Sciences\\
$^5$Microsoft Research Asia\\
\emails
\{visca.panda.xu, is.fenghao, yuguoxin98\}@gmail.com,
min.yang@siat.ac.com,
xitwan@microsoft.com,
aoxiang@ict.ac.cn
}

\begin{document}

\maketitle

\begin{abstract}

\footnotetext[6]{ Xu and Feng make equal contributions to this work.}
\footnotetext[7]{Xiang is the corresponding author.}

Most recent existing aspect-term level sentiment analysis~(ATSA) approaches combined various neural network models with delicately carved attention mechanisms built upon given aspect and context to generate refined sentence representations for better predictions. 
In these methods, aspect terms are always provided in both training and testing process which may degrade aspect-level analysis into sentence-level prediction. 
However, the annotated aspect term might be unavailable in real-world scenarios which may challenge applicability of the existing methods. 
In this paper, we aim to improve ATSA by discovering the potential aspect terms of the predicted sentiment polarity when the aspect terms of a test sentence are unknown. We access this goal by proposing a capsule network based model named CAPSAR. In CAPSAR, sentiment categories are denoted by capsules and aspect term information is injected into sentiment capsules through a sentiment-aspect reconstruction procedure during the training. As a result, coherent patterns between aspects and sentimental expressions are encapsulated by these sentiment capsules. 
Experiments on three widely used benchmarks demonstrate these patterns have potential in exploring aspect terms from test sentence when only feeding the sentence to the model. Meanwhile, the proposed CAPSAR can clearly outperform SOTA methods in standard ATSA tasks.

\end{abstract}

\section{Introduction}\label{sec:intro}

Aspect-level sentiment analysis is an essential building blocks of sentiment analysis~\cite{liu2012sentiment}.
It aims at extracting and summarizing the sentiment polarities of given aspects of entities, i.e. \emph{targets}, from customers' comments. Two subtasks are explored in this field, namely Aspect-Term level Sentiment Analysis~(ATSA) and Aspect-Category level Sentiment Analysis~(ACSA). 
The purpose of ATSA is to predict the sentiment polarity with respect to given targets appearing in the text. 
For example, consider the sentence ``\emph{The camera of iPhone XI is delicate, but it is extremely expensive.}'', ATSA may ask the sentiment polarity towards the given target ``\emph{camera}''.
Meanwhile, ACSA attempts to predict the sentiment tendency regarding a given target chosen from predefined categories, which may not explicitly appear in the comments.
Take the same sentence as an example, ACSA asks the sentiment towards the aspect ``\emph{Price}'' and derives a negative answer. 
In this paper, we aim at addressing the ATSA task.

Conventional approaches~\cite{hu2004mining,ding2008holistic,kiritchenko2014nrc} incorporated linguistic knowledge, such as sentiment-lexicon, syntactic parser, and negation words, etc., and tedious feature engineering into the models to facilitate the prediction accuracy.
Recently, supervised deep neural networks, e.g. Recurrent Neural Network~(RNN)~\cite{tang2016effective,zhang2016gated}, Convolution Neural Network~(CNN)~\cite{xue2018aspect,li2018transformation} and attention mechanism~\cite{tang2016aspect,wang2016attention,ma2017interactive,chen2017recurrent,huang2018aspect,fan2018multi,wang2018learning,luo2019unsupervised,mao2019aspect} have shown remarkable successes without cumbersome feature designing. 
These models are able to effectively screen unrelated text spans and detect the sentiment context about the given target.

Despite these efforts, there is still a major deficiency in previous deep neural network based studies. Specifically, fully labeled aspect terms and their locations in sentence are explicitly required in both training and test process for recent methods, which may derive them degrade to sentence-level prediction and would fail for the test data without such annotations. 
To acquire aspect terms on predicted sentences, automatic aspect term detection may lead to error accumulation~\cite{wang2019www}, and manually identifying is inefficient even infeasible. 
To support more authentic applications, it calls for an approach that is able to predict potential aspect-related sentiments based on the sentence and what it has learned from the training set.


To this end, we propose a capsule network-based approach to remedy the above problem. 
Compared with previous studies, our method is able to explore featured sentiments so as to answer the question: ``\emph{What are the protagonists of the predicted sentiment polarity?}'' 
We access this goal by leveraging the capsule network\footnote{Here we refer to the capsule network proposed by~\cite{sabour2017dynamic}. Though the models in~\cite{wang2018www} and \cite{wang2019www} also called capsule network in their papers, they are basically built upon RNN and attention mechanisms with distinct concepts and implementations.}~\cite{sabour2017dynamic}, 
which has achieved promising results in computer vision~\cite{sabour2017dynamic,hinton2018matrix}, natural language processing~\cite{zhao2018investigating,zhang2018multi,zhang2018attention,Jiang2019dataset,du2019capsule} and recommendation tasks~\cite{Li2019sigir}, etc. 

The core idea of capsule network is the unit named \emph{capsule}, which consists of a group of neurons in which its activity vector can represent the instantiation parameters of a specific type of entity. 
The length of activity vector denotes the probability that the entity exists and its orientation can encode the properties of the entity.
Inspired by that, we propose CAPSAR~(\textbf{CAP}sule network with \textbf{S}entiment-\textbf{A}spect \textbf{R}econstruction) framework by leveraging capsules to denote sentiment categories and enforce the potential aspect information as the corresponding properties.

Specifically, during the training process, the sentence is first encoded with given aspects through a location proximity distillation.
Then the encoded sentence representations are fed to hierarchical capsule layers and the final capsule layer represents all the concerned sentiment categories. 
To capture coherent patterns among aspects and sentimental expressions, the sentiment capsules are encouraged to encode the information about the aspect terms. We implement such procedure by reconstructing the aspect with the sentiment capsules. 
The reconstruction loss is taken as an additional regularization during the training. 
During the test phase, if the annotated aspect term is unseen by the model, CAPSAR can also make prediction and the potential aspect terms in the sentence could be detected by de-capsulizing the sentiment capsules. 
We evaluate the proposed methods on three widely used benchmarks. The results show the model has potential in unearthing aspect terms for new sentences, and it can also surpass SOTA baselines in standard aspect-term level sentiment analysis tasks. 


    

\section{Related Work}\label{sec:survey}

\noindent\textbf{Sentiment analysis based on neural network.}
Neural network approaches have achieved promising results on both document level~\cite{luo2018beyond,Yang2016HierarchicalAN,DBLP:conf/ijcai/WangWZY17} and sentence level~\cite{socher2011semi} sentiment classification tasks without expensive feature engineering. 
\cite{wang2018www} firstly adopted capsules into document-level sentiment analysis, but their capsule is still based on RNN and attentions, which is different with the capsule designs in~\cite{sabour2017dynamic}.

\noindent\textbf{Aspect level sentiment classification.}
Aspect level sentiment classification is an emerging essential research topic in the field of sentiment analysis. 
The purpose is to infer the polarity with respect to aspect phrase or predefined aspect categories within the text. 
\cite{tang2016effective,zhang2016gated} used multiple RNN layers to jointly model the relations between target terms and their left and right context. Attention-based methods was brought to this field by many researches~\cite{tang2016aspect,wang2016attention,ma2017interactive,chen2017recurrent,huang2018aspect,fan2018multi} and have achieved promising results. 
However, the final representation may still fail to capture the accurate sentiment due to target-sensitive problem~\cite{wang2018target} or because of the noise in data. 
Recently, \cite{wang2019www} proposed to use capsules to perform aspect-category level sentiment analysis. However as their previous work~\cite{wang2018www}, the basic capsule module is based on attention mechanisms, which is entirely different with ours. 
\section{The CAPSAR Model}\label{sec:model}

\subsection{Model Overview}

\begin{figure}[!htbp]
    \centering
    \includegraphics[height=7.5cm, width=8.2cm]{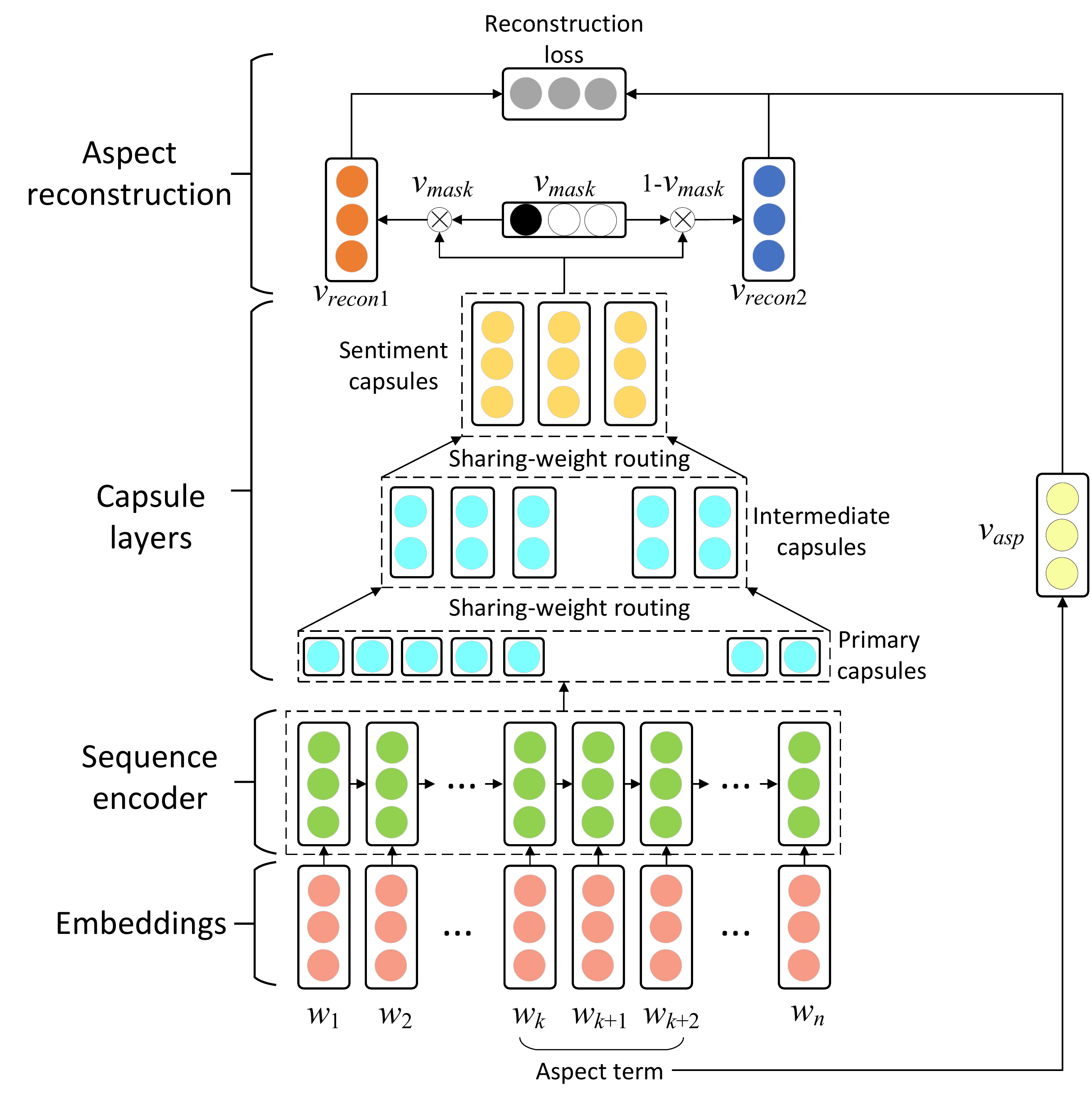}
    \caption{The network architecture of CAPSAR. }
    \label{fig:model}
\end{figure}

The overall architecture of CAPSAR is shown in Figure~\ref{fig:model}. 
It starts from an embedding representation of words. In particular, we represent the $i$-th sentence in a dataset $\mathbb{D}$ with $m$ sentences as $\{w_1^{(i)},w_2^{(i)},...,w_{n_i}^{(i)}\}$, where $i\in [1,...,m]$, $n_i$ is the sentence length, and $w_{.}^{(i)}\in W$ denotes a word where $W$ is the set of vocabulary. 
The embedding layer encodes each word $w_{t}^{(i)}$ into a real-value word vector $x_t^{(i)} \in \mathbb{R}^{D_x}$ from a matrix $M\in\mathbb{R}^{|W|\times D_x}$, where $|W|$ is the vocabulary size and $D_x$ is the dimension of word vectors.
The sentence is encoded by a sequence encoder to construct a sentence representation. Next, the output of the sequence encoder is fed to 3-layer capsules. The up-most capsule layer contains $C$ \emph{sentiment capsules}, where $C$ is the number of sentiment categories. The capsules layers are communicated with a simple yet effective \emph{sharing-weight} routing algorithm.

During training, one objective of our model is to maximize the length of sentiment capsules corresponding to the ground truth since it indicates the likelihood of potential sentiments. 
Meanwhile, these active vectors of the sentiment capsules are used to model the connections between the considered aspect terms and its corresponding sentiment via an aspect reconstruction. The distance between the reconstructed aspect representation produced by sentiment capsules and the given aspect embedding\footnote{The aspect embedding is calculated by the average of the word embeddings that form the aspect term.} is regarded as an additional regularization. In this manner, we encourage the sentiment capsules to learn the aspect information as their active vectors' orientations. 
In the test process, a sentiment capsule will be ``active'' if its length is above a user-specific threshold, e.g. $0.5$. All others will then be ``inactive''. The sentiment prediction of a test sentence will be determined by the sentiment category associated with the active sentiment capsules.

\subsection{Sequence Encoder}

In our model, we adopt Bi-GRU as the sequence encoder for simplicity. 
For $i$-th sentence at step $t$, the corresponding hidden state $h_t^{(i)}$ are updated as follows.

\begin{equation}\vspace{-0.5mm}
h_{t}^{(i)}=\begin{bmatrix}\overrightarrow{h}_{t}^{(i)}\\\overleftarrow{h}_{t}^{(i)}\end{bmatrix}
=\begin{bmatrix}\overrightarrow{\mbox{GRU}}(x_{t}^{(i)})\\\overleftarrow{\mbox{GRU}}(x_{t}^{(i)})\end{bmatrix},\quad t=1,\dots,n_i,
\end{equation}

where $h_{t}^{(i)}$ concatenates hidden states of the $t$-th word in the $i$-th sentence from both directions.
Note that more advanced encoders such as LSTM~\cite{hochreiter1997long} or BERT~\cite{devlin2018bert}, can also be utilized as the sequence encoder. We will introduce how to combine CAPSAR with BERT in the following part.

\subsection{Location Proximity with Given Aspect}

In order to highlight potential opinion words that are closer to given aspect terms, we adopt a location proximity strategy, which is observed effective in~\cite{chen2017recurrent,li2018transformation}. 
Specifically, we calculate relevance $l_t^{(i)}$ between the $t$-word and the aspect\footnote{$t$ is possibly larger than $n_i$ because of sentence padding.}.

\begin{equation}\label{eq:location}\small
l_t^{(i)}=\left\{
\begin{array}{lcl}
1+\max(0, \alpha+n_i/\beta-|\gamma*(k-t)|) & & {t \leq n_i}\\
0 & & {t>n_i}
\end{array} \right.
\end{equation}
where $k$ is the index of the first aspect word, $n_i$ is the sentence length, $\alpha$, $\beta$ and $\gamma$ are pre-specified constants. 


We use $l$ to help the sequence encoder locate possible key words w.r.t the given aspect.
\begin{equation}\label{eq:hhat}\small
\hat{h}_t^{(i)}=h_t^{(i)}*l_t^{(i)}, t\in[1, n_i], i\in[1,m]
\end{equation}

Based on Eq.~\ref{eq:location} and \ref{eq:hhat}, the salience of words that are distant to the aspect terms will be declined. 
Note that such location proximity could be optional in the test process when the annotated aspect terms are unavailable.

\subsection{Capsule Layers with Sharing-weight Routing}

The capsule layers of CAPSAR consist of a primary capsule layer, an intermediate capsule layer and a sentiment capsule layer. 
The primary capsule layer contains a group of neurons which are constructed by the hidden vectors of the sequence encoder. Specifically, we simply perform convolutional operation over $h_{n_i}^{(i)}$ for $i$-th sentence and take its output to formulate the primary capsules. As a result, the primary capsules may contain the sentence coupled with aspect representations. 

Next, the primary capsules are transformed into the intermediate layer and the subsequent sentiment capsule layer via a \emph{sharing-weight} routing mechanisms. Unlike the conventional dynamic routing algorithm in~\cite{sabour2017dynamic}, our routing algorithm simultaneously keeps local-proximity information and significantly reduces training parameters.


The sharing-weight routing algorithm shares the weights between different children and the same parent. 
Specifically, we denote the output vector of a capsule $i$ at level $L$ and the total input vector of a capsule $j$ at level $L+1$ as ${p}_i\in\mathbb{R}^{D_{L}}$ and ${\tilde{q}}_j\in\mathbb{R}^{D_{L+1}}$, respectively. 
We use a unified transformation weight matrix ${W}_j \in \mathbb{R}^{D_{L+1}\times D_{L}}$ for the capsule $j$ at level $L+1$ to compute the prediction vectors, i.e. ${\hat{p}}_{j|i}\in\mathbb{R}^{D_{L+1}}$, for every possible child capsule $i$ at level $L$.
As a result, the total input ${\tilde{q}}_j$ of capsule $j$ is updated as
\begin{equation}\label{eq:newpi}\small
{\tilde{q}}_j = \sum_{i}c_{ij}\hat{p}_{j|i}, \ \ {\hat{p}}_{j|i}={W}_{j}{p}_i
\end{equation}
where $c_{ij}$ denotes coupling coefficients between capsule $i$ and $j$ and is initialized with equal probability. 
During the iterative dynamic routing process, $c_{ij}$ is updated to $q_j\cdot\hat{p}_{j|i}$, where $q_j$ is the output vector of capsule $j$ computed by the \texttt{squash} function. 
\begin{equation}\small
q_j = \frac{\vert\vert \tilde{q}_j \vert\vert^2}{1 + \vert\vert \tilde{q}_j \vert\vert^2}\frac{\tilde{q}_j}{\vert\vert \tilde{q}_j \vert\vert}
\end{equation}


Compared with the conventional dynamic routing algorithm, the sharing-weight routing algorithm clearly reduces the number of parameters and saves computational cost. 
For example, if two consecutive capsule layers have $M$ and $N$ capsules and the dimensions are $D_{L}$ and $D_{L+1}$ respectively, then the number of parameters to be learned in this layer will be reduced by $(M-1) \times N \times D_{L} \times D_{L+1}$ compared with the original routing algorithm in~\cite{sabour2017dynamic}.


\subsection{Model Training with Aspect Reconstruction}
The training objective of this model is two-fold. 
On one hand, we aim to maximize the length of the correct sentiment capsules since it indicates the probability that the corresponding sentiment exists.
To this end, we use a margin loss for every given sentence $i$
\begin{equation}\small
\begin{aligned}
L_1^{(i)} &= v_\mathit{{mask}}^{(i)}\max(0,m^+-\vert\vert v_{\mathit{prob}}^{(i)}\vert\vert)^2\\
&+(1-v_{\mathit{mask}}^{(i)})\max(0,\vert\vert v_{\mathit{prob}}^{(i)}\vert\vert-m^-)^2
\end{aligned}
\end{equation}

Here $v_{\mathit{prob}}^{(i)}=(||q_{1}^{(i)}||,\cdots,||q_{C}^{(i)}||)$ where $q_{j}^{(i)}$ denotes the output vector of sentiment capsule $j$ for the sentence $i$. 
Each element in such $v_{\mathit{prob}}^{(i)}$ indicates the existence probability of the corresponding sentiment in sentence $i$; $v_\mathit{{mask}}^{(i)}$ is the mask for sentence $i$; $m^+$ and $m^-$ are hyper-parameters.

On the other hand, we attempt to encourage the sentiment capsules to capture interactive patterns between the aspect and their corresponding sentiments. 
To this end, we utilize the output vectors of all the sentiment capsules $q_j$($j\in [1, C]$) to participate in reconstructing the representation of aspect terms. 
Specifically, suppose $v_{\mathit{mask}}$ is a one-hot mask\footnote{The dimension of $v_{mask}$ is $C$.} whose element representing the ground truth sentiment is 1, and the rest values are 0.
Then we derive two vectors, namely $v_{\mathit{recon1}}$ and $v_{\mathit{recon2}}$, through this mask and the sentiment capsules.
First, we mask out all but the output vector of the correct sentiment capsule by $v_{\mathit{mask}}$.
Then $v_{\mathit{recon1}}$ is derived by transforming the masked output vector through a fully-connect layer.
$v_{\mathit{recon2}}$ can be derived in a similar manner where $1-v_{\mathit{mask}}$ is used as the mask.
We force both $v_{\mathit{recon1}}$ and $v_{\mathit{recon2}}$ have the same dimension of word embedding, i.e., $v_{\mathit{recon1}}$, $v_{\mathit{recon2}}$ $\in \mathbb{R}^{D_{x}}$, and they are contributed to the aspect reconstruction during the training. 

Suppose a given aspect embedding\footnote{If there are more than one aspect in a same sentence, every aspect will be separately trained.} of the sentence $i$ is denoted as $v_{\mathit{asp}}^{(i)}$, another training objective in our CAPSAR is to minimize the distance between $v_{\mathit{asp}}$ and $v_{\mathit{recon1}}$, and to maximize that between $v_{\mathit{asp}}$ and $v_{\mathit{recon2}}$.
\begin{equation}\small
L_2^{(i)} = -{v_{\mathit{asp}}^{(i)}}\frac{v_{recon1}^{(i)}}{\vert\vert v_{recon1}^{(i)}\vert\vert}+ {v_{\mathit{asp}}^{(i)}}\frac{v_{recon2}^{(i)}}{\vert\vert v_{recon2}^{(i)}\vert\vert}
\end{equation}

Finally, the overall loss is the combination of $L_1^{(i)}$ and $L_2^{(i)}$, and a hyper-parameter $\lambda$ is used to adjust the weight of $L_2$.
\begin{equation}\label{eq:loss}\small
Loss = \sum_{i}(L_1^{(i)} + \lambda L_2^{(i)})
\end{equation}\vspace{-3mm}

For prediction, a sentence and an optional aspect in the sentence are fed to the network and the polarity attached to the sentiment capsule with the largest length will be assigned. 

\subsection{Combining CAPSAR with BERT}
Our CAPSAR is meanwhile easily extended that utilizes the features learnt from large-scale pre-trained encoders, e.g. BERT~\cite{devlin2018bert}. An upgraded model,namely CAPSAR-BERT, is achieved by replacing the sequence encoder with BERT in CAPSAR. The other structures are kept the same. In this manner, the strength of BERT and the proposed structures could be combined.

\section{Experiments}\label{sec:expt}
In this section, we verify the effectiveness of CAPSAR. 
Firstly, we verify the ability of CAPSAR on perceiving the potential aspect terms when they are unknown.
Secondly, we investigate the performance of CAPSAR on standard ATSA tasks, where the aspect terms are always known for test sentences. Finally, we demonstrate the detailed differences of compared methods by case studies.
\subsection{Datasets}
Three widely used benchmark datasets are adopted in the experiment whose statistics are shown in Table~\ref{tab:statistics}. \textbf{Restaurant} and \textbf{Laptop} are from SemEval2014 Task 4\footnote{\url{http://alt.qcri.org/semeval2014/task4/index.php?id=data-and-tools}}, which contain reviews from Restaurant and Laptop domains, respectively. 
We delete a tiny amount of data with conflict labels which follows previous works~\cite{wang2016attention,tang2016aspect}. 
\textbf{Twitter} is collected by~\cite{dong2014adaptive} containing twitter posts. Though these three benchmarks are not large-scale datasets, they are most popular and fair test beds for recent methods.

\begin{table}[!htbp]
    \centering
    \scriptsize
        \begin{tabular}{ccccc}
        \toprule
        \textbf{Dataset} &       & \textbf{Neg.} & \textbf{Neu.} & \textbf{Pos.} \\
        \midrule
        \multirow{2}[1]{*}{\textbf{Restaurant}} & \textbf{Train} & 807   & 637   & 2164 \\
        & \textbf{Test} & 196   & 196   & 728 \\
        \hline
        \multirow{2}[0]{*}{\textbf{Laptop}} & \textbf{Train} & 870   & 464   & 994 \\
        & \textbf{Test} & 128   & 169   & 341 \\
       \hline
        \multirow{2}[1]{*}{\textbf{Twitter}} & \textbf{Train} & 1562  & 3124  & 1562 \\
        & \textbf{Test} & 173   & 346   & 173 \\
        \bottomrule
    \end{tabular}%
    \vspace{-2mm}
    \caption{Statistics of datasets.}
      \label{tab:statistics}\vspace{-5mm}
\end{table}%

\begin{table*}[!htbp]\scriptsize
    \centering
    \begin{tabular}{cccccccc}
        \toprule
        & &  \multicolumn{2}{c}{\textbf{Restaurant}} & \multicolumn{2}{c}{\textbf{Laptop}} & \multicolumn{2}{c}{\textbf{Twitter}} \\ \cline{3-8} 
        & \textbf{Models} & \textbf{Accuracy} & \textbf{Macro-F1} & \textbf{Accuracy} & \textbf{Macro-F1} & \textbf{Accuracy} & \textbf{Macro-F1} \\
        \midrule
        \multirow{8}{*}{\textbf{Baselines}}
        &{ATAE-LSTM} & {0.7720} &  {NA} & {0.6870} & {NA} & {NA} & {NA} \\
        &{TD-LSTM} & {0.7560} & {NA} & {0.6810} & {NA} & {0.6662$^{*}$} & {0.6401$^{*}$} \\
        &{IAN} & {0.7860} &{NA} &{0.7210} &{NA} &{NA} &{NA}\\
        &{MemNet(3)} & {0.8032} &{NA} &{0.7237} &{NA} &{0.6850$^{*}$} &{0.6691$^{*}$} \\
        &{RAM(3)} & {0.8023} & {0.7080} & {0.7449} & {0.7135} & {0.6936} & {0.6730} \\
        &{MGAN} & 0.8125 & 0.7194 & 0.7539  & \textbf{0.7247} &  0.7254 & 0.7081 \\
        &{ANTM} & {0.8143} & {0.7120} & {0.7491} & {0.7142} & {0.7011} & {0.6814} \\

        \midrule
        \multirow{3}{*}{\textbf{Ablation Test}}
        &{CAPSAR w/o R} & 0.8185 & 0.7216 &0.7484 & {0.7039} & 0.7255 & 0.7067 \\
        &{CAPSAR w/o H} &0.8188  &0.7226  &0.7461 & {0.7054}  &0.7298  &0.7080\\ 
        &{CAPSAR} & \textbf{0.8286}& \textbf{0.7432} & {\textbf{0.7593}}  &{0.7221}& {\textbf{0.7368}} & {\textbf{0.7231}} \\
        
        \midrule
        \multirow{2}{*}{\textbf{Combine BERT}}
        &{BERT} & 0.8476 & 0.7713&0.7787 &0.7371 & 0.7537 & 0.7383 \\
        &{CAPSAR-BERT} & \textbf{0.8594} & \textbf{0.7867} &\textbf{0.7874} &\textbf{0.7479} & \textbf{0.7630} & \textbf{0.7536} \\
        
        \bottomrule
    \end{tabular}%
    \vspace{-2mm}
    \caption{The average accuracy and F1-score on standard ATSA tasks. The results with `*' are retrieved from the papers of RAM, and other results of baselines are rerieved from corresponding papers.}
    \label{tab:results}%
    \vspace{-3mm}
\end{table*}%

\subsection{Compared Methods}
We compare our method with several SOTA approaches.

\textbf{ATAE-LSTM}~\cite{wang2016attention} appends aspect embedding with each input word embeddings.
\textbf{TD-LSTM}~\cite{tang2016effective}
employs two LSTMs to model contexts of the targets and performs predictions based on the concatenated context representations.
\textbf{IAN}~\cite{ma2017interactive} interactively learns the context and target representation.
\textbf{MemNet}~\cite{tang2016aspect} feds target word embedding to multi-hop memory to learn better representations. 
\textbf{RAM}~\cite{chen2017recurrent}
uses recurrent attention to capture key information on a customized memory. 
\textbf{MGAN~\cite{fan2018multi}} equips a multi-grained attention to address the aspect has multiple words or larger context. 
\textbf{ANTM~\cite{mao2019aspect}} adopts attentive neural turing machines to learn dependable correlation of aspects to context.
\textbf{CAPSAR} and \textbf{CAPSAR-BERT} are models proposed in this paper. 
\textbf{BERT}~\cite{devlin2018bert} is also compared to show the improvement of CAPSAR-BERT.

\subsection{Experimental Settings}

In our experiments, we implement our method by Keras 2.2.4. 
The word embedding is initialized by Glove 42B~\cite{pennington2014glove} with dimension of 300. 
The max length for each sentence is set to 75. 
The batch size of training is 64 for 80 epochs, and Adam~\cite{kingma2014adam} with default setting is taken as the optimizer. 
For sequence encoder, we adopt the Bi-GRU with the dropout rate of 0.5 for CAPSAR. The pre-trained BERT of the dimension of 768 is used in CAPSAR-BERT. For hyper-parameters in Eq.~\ref{eq:location}, $\alpha$, $\beta$, and $\gamma$ are set to be 3, 10 and 1. 
For the capsule layers, there are 450 primary capsules with dimensions of 50, 30 intermediate capsules with dimensions of 150 and 3 sentiment capsules with dimensions of 300. 
The default routing number is 3.
For hyper-parameters in the loss function~(cf. Eq.~\ref{eq:loss}), we set $m^+$, $m^-$, $\lambda$ to be 1.0, 0.1 and 0.003 respectively. 
Evaluation metrics we adopt are Accuracy and Macro-F1, and the latter is widely used for recent ATSA tasks since it is more appropriate for datasets with unbalanced classes.

\subsection{Results on Standard ATSA}
\subsubsection{Main Results}
Table~\ref{tab:results} demonstrates the performances of compared methods over the three datasets on the standard ASTA tasks. 
On such setting, every aspect term is known to all the models, and each model predicts the corresponding polarity for a given aspect term. Here we only consider the longest sentiment capsule is active. 
All the reported values of our methods are the average of 5 runs to eliminate the fluctuates with different random initialization, and the performance of baselines are retrieved from their papers for fair comparisons. The best performances are demonstrated in bold face. 
From the table, we observe CAPSAR has clear advantages over baselines on all datasets. Our model can outperform all the baselines by a large-margin on both evaluate measures except the F1 on Laptop dataset.
Meanwhile, we also observe that CAPSAR-BERT further improves the performance of BERT. It demonstrates the advantages by combining CAPSAR with advanced pre-trained model.


\begin{table*}[!htbp]
    \centering
    \scriptsize
    \begin{tabular}{p{20em}cccc}
        \toprule
        \multicolumn{1}{c}{\textbf{Sentence}} & \textbf{CAPSAR} & \textbf{ANTM} & \textbf{MGAN} & \textbf{RAM} \\
        \midrule
        1. The \textcolor[rgb]{0,0,1}{[{\bfseries chocolate raspberry cake}]$\rm_{Pos}$ is heavenly} - not too sweet , but full of \textcolor[rgb]{1,0,0}{[{\bfseries flavor}]$\rm_{Pos}$.} & (\textcolor[rgb]{0,0,1}{Pos},\textcolor[rgb]{1,0,0}{Pos}) & (\textcolor[rgb]{0,0,1}{Pos},\textcolor[rgb]{1,0,0}{Neg}\ding{55}) & (\textcolor[rgb]{0,0,1}{Neg}\ding{55},\textcolor[rgb]{1,0,0}{Pos}) & (\textcolor[rgb]{0,0,1}{Pos},\textcolor[rgb]{1,0,0}{Neg}\ding{55}) \\
        \midrule
        2. Not only was the sushi fresh , they also served other \textcolor[rgb]{0,0,1}{[{\bfseries entrees}]$\rm_{Neu}$} allowed each guest something to choose from and \textcolor[rgb]{1,0,0}{we all left happy $($try the [{\bfseries duck}]$\rm_{Pos}$!} & (\textcolor[rgb]{0,0,1}{Neu},\textcolor[rgb]{1,0,0}{Pos}) & (\textcolor[rgb]{0,0,1}{Pos}\ding{55},\textcolor[rgb]{1,0,0}{Pos}) & (\textcolor[rgb]{0,0,1}{Pos}\ding{55},\textcolor[rgb]{1,0,0}{Pos}) & (\textcolor[rgb]{0,0,1}{Pos}\ding{55},\textcolor[rgb]{1,0,0}{Pos}) \\
        \midrule
        3. The \textcolor[rgb]{0,0,1}{[{\bfseries baterry}]$\rm_{Pos}$} is very \textcolor[rgb]{0,0,1}{longer} . & \textcolor[rgb]{ .000,  .000,  1}{Pos} & \textcolor[rgb]{ .0,  .0,  1}{Neg}\ding{55} & \textcolor[rgb]{  .000,  .000,  1}{Pos} & \textcolor[rgb]{ .0,  .0,  1}{Neg}\ding{55} \\
        \midrule
        4. \textcolor[rgb]{0,0,1}{[{\bfseries Startup times}]$\rm_{Neg}$} are incredibly \textcolor[rgb]{0,0,1}{long} : over two minutes.  & \textcolor[rgb]{ .0,  .0,  1}{Neg} & \textcolor[rgb]{ .0,  .0,  1}{Neg} & \textcolor[rgb]{ .0,  .0,  1}{Pos}\ding{55} & \textcolor[rgb]{ .0,  .0,  1}{Pos}\ding{55} \\
        \midrule
        5. However I chose two day \textcolor[rgb]{0,0,1}{[{\bfseries shipping}]$\rm_{Neg}$} and it took over a week to arrive. & \textcolor[rgb]{ .0,  .0,  1}{Neu}\ding{55}& \textcolor[rgb]{ .0,  .0,  1}{Neu}\ding{55}& \textcolor[rgb]{ .0,  .0,  1}{Pos}\ding{55} & \textcolor[rgb]{ .0,  .0,  1}{Pos}\ding{55} \\
        \bottomrule
    \end{tabular}%
\vspace{-2mm}
    \caption{Prediction examples of some of the compared methods. The abbreviations Pos, Neu and Neg in the table represent positive, neutral and negative. \ding{55} indicates incorrect prediction.}
\label{tab:case}%
\vspace{-2mm}
\end{table*}%

\subsubsection{Ablation Test}
Next we perform ablation test to show the effectiveness of  component of CAPSAR. We remove the aspect reconstruction and intermediate capsule layer, respectively, and derive two degrade models, namely \textbf{CAPSAR w/o R} and \textbf{CAPSAR w/o H}. Their performance on the three datasets are exhibited in Table~\ref{tab:results}, and we can observe the degraded model gives a clear weaker performance than CAPSAR, without the sentiment-aspect reconstruction. 
We conjecture such a regularizer might be able to learn the interactive patterns among the aspects with the complex sentimental expressions. Meanwhile, we also observe the advantage of the intermediate capsule layer, which illustrates the hierarchical capsule layers may be stronger in learning aspect-level sentiment features.   

\subsubsection{Case Studies}

Then we illustrate several case studies on the results of ATSA task in Table~\ref{tab:case}. The predictions of CAPSAR, ANTM, MGAN and RAM are exhibited. We directly run their available models on test set to get the results. The input aspect terms are placed in the brackets with their true polarity labels as subscripts. 
``Pos'', ``Neu'' and ``Neg'' in the table represent positive, neutral and negative, respectively.
Different targets are demonstrated by different colors, such as blue and red, etc.
The context which may support the sentiment of targets is manually annotated and dyed with the corresponding color.

For instance, in the first sentence, one target is ``\emph{chocolate raspberry cake}'' of which the sentiment is positive and the context ``\emph{is heavenly}'' supports this sentiment. 
In this case, the results of CAPSAR, ANTM, and RAM are correct. For the second aspect ``\emph{flavor}'', our method and MGAN predict correctly.
We conjecture the reason is the sentence contains turns in its expression, which may confuse the existing neural network models. 
Similar observation is achieved on the second sentence in which only our method can predict different sentiments for distinct aspects.



Next we discuss a target-sensitive case which is shown by the third and the fourth sentences in the same table. The word ``long'' in the exhibited two sentences indicates entirely opposite sentiment polarities, since the expressed sentiment also depends on the considered aspects. These cases are challenge for algorithms to identify the sentiment of each sentence correctly. Among the demonstrated methods, only our approach can predict them all successfully. ANTM, as a strong competitor, can predict one of them correctly. The others fail to give any correct prediction. We do not claim our CAPSAR can perfectly address the target-sensitive cases, however, the results give an initial encouraging potential. We argue that it is the sentiment-aspect reconstruction in our model which makes aspect and its corresponding sentiment become more coupled, and it might be one of the essential reasons why our model achieves a salient improvement on ATSA task. How to specifically explore capsule networks for target-sensitive sentiment analysis is out of the scope of this paper.


\begin{table}[!tphb]\scriptsize
    \vspace{-0.2cm}
    \setlength{\belowcaptionskip}{-0.2cm}
    \centering
    \begin{tabular}{p{1cm}<{\centering}p{1cm}<{\centering}p{1cm}<{\centering}p{1cm}<{\centering}p{1cm}<{\centering}p{1cm}<{\centering}}
        \toprule
        \multirow{2}{*}{\textbf{Datasets}} & \textbf{Avg. Aspect}&\textbf{Avg. SenLen}&\textbf{Pre.@1}& \textbf{Rec.@5}&\textbf{mAP}    \\
        \midrule
        \textbf{Resturant} & 2.76 &16.25&0.8233 & 0.7884&0.7139\\ 
        \textbf{Laptop} &  2.54 &15.79&0.6408&0.7557&0.6173\\ 
        \bottomrule
    \end{tabular}%
    \caption{The average Precision@1, Recall@5 and mAP on aspect term detection. The column “Avg. Aspect” and “Avg.SenLen” indicate average number of words on aspect terms in each sentence and average length of each sentence on the test set,  respectively.}
    \label{tab:aspectdetection}%
    \vspace{-2mm}
\end{table}%


For error analysis, we find that all the listed models cannot predict correctly on the last sentence of Table~\ref{tab:case}. 
By looking closer to this sentence, we recognize that the sentiment polarity of this sentence comes from implicit semantics instead of its explicit opinion words.
It indicates implicit semantics inference behind sentences is still a major challenge of neural network models, even exploiting capsule networks.

\subsection{Results on Aspect Term Detection}\label{sec:reconstruction}

Next, we investigate whether CAPSAR can detect potential aspect terms when they are unknown during the test. To this end, we use the trained model to predict every test sentence on Restaurant and Laptop datasets but we intentionally conceal the information about the aspect terms in the input. In another word, the model is only fed the test sentence without any other additional input. 
Then we de-capsulize the sentiment capsule whose length is longer than $0.5$ and compute normalized dot-product between its reconstructed vector and every word embedding in the test sentence. 
These dot-products can be regarded as the probabilities representing the possibility of the word to be part of an aspect term.

A sentence may simultaneously contain multiple sentiments~(c.f. the $2$nd sentence in Table~\ref{tab:case}), which derives more than one sentiment capsule whose length surpasses $0.5$. As a result, we detect the potential aspect terms for every active sentiment capsule, respectively, in our evaluation. 
There could be more than one aspect terms for each sentiment category in the same sentence~(c.f. the $1$st sentence in Table~\ref{tab:case}). 
Hence we compute Precision@$k$, Recall@$k$ and mean Average Precision~(mAP) to comprehensively verify the effectiveness of CAPSAR on aspect term detection.

\begin{figure}[!htbp]
    \vspace{-0.2cm}
    \setlength{\belowcaptionskip}{-0.2cm}
    \centering
    \includegraphics[height=2.88cm,width=8.5cm]{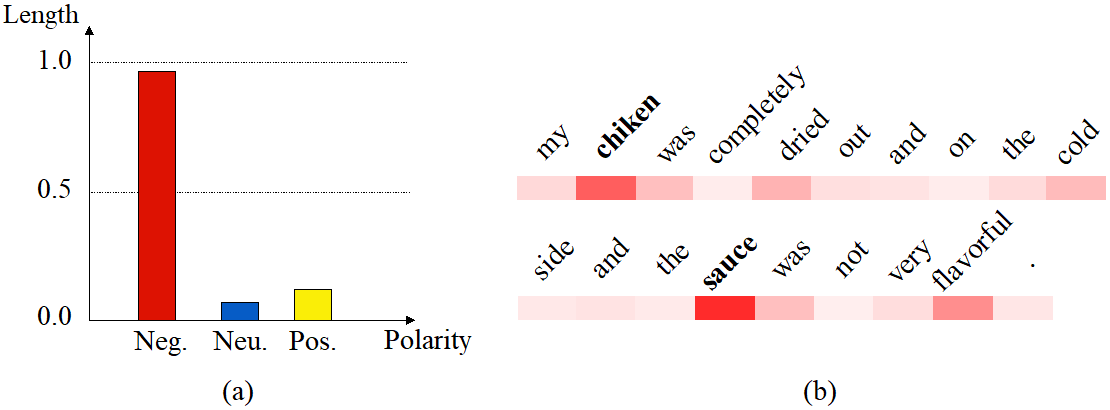}\vspace{-2mm}
    \caption{The visualization of aspect term detection when single sentiment capsule is active. The real aspect terms are marked in bold face in the sub-figure~(b).}
    \label{fig:aspectdetectcase1}
\end{figure}

\begin{figure}[!htbp]
    \vspace{-0.2cm}
    \setlength{\belowcaptionskip}{-0.2cm}
    \centering
    \includegraphics[height=2.88cm,width=8.5cm]{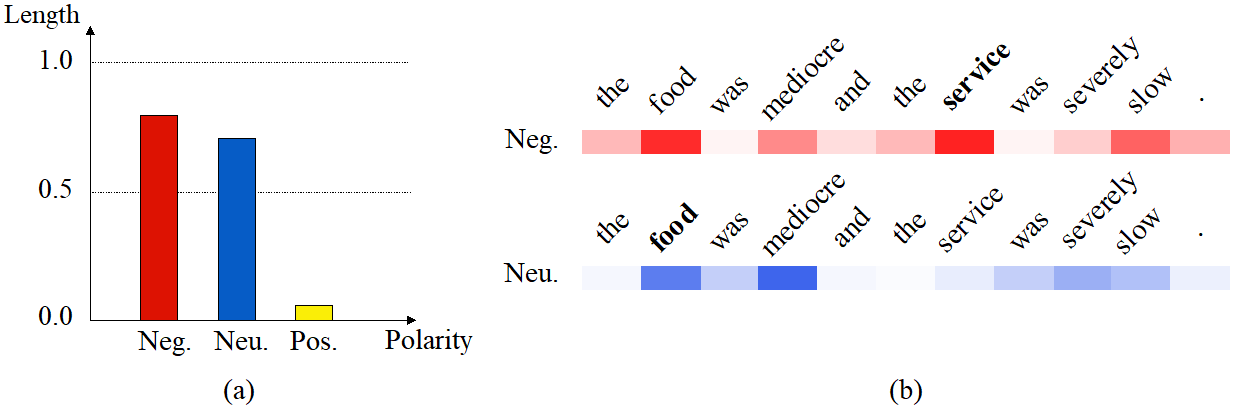}\vspace{-2mm}
    \caption{The visualization of aspect term detection when multiple sentiment capsules are active. The real aspect terms are marked in bold face in the sub-figure~(b), respectively. }\label{fig:aspectdetectioncase2}
\end{figure}


We retrain our CAPSAR five times and use the trained model to detect aspect terms on the test set. 
Table~\ref{tab:aspectdetection} shows the average results. From the table we observe that CAPSAR shows an encouraging ability to extract aspect terms even though they are unknown in new sentences. Meanwhile, the model achieves better performance on Restaurant dataset. We conjecture the reason is the Laptop dataset has more complicated aspect terms such as ``Windows 7''. 

Finally, we visualize two test sentences in this evaluation shown as Figure~\ref{fig:aspectdetectcase1} and \ref{fig:aspectdetectioncase2}. Figure~\ref{fig:aspectdetectcase1} demonstrates a case that only one sentiment capsule is active. The sentiment capsule length is exhibited in Fig.~\ref{fig:aspectdetectcase1}(a), and the corresponding dot-products are demonstrated as Fig.~\ref{fig:aspectdetectcase1}~(b). The darker the color in Fig.~\ref{fig:aspectdetectcase1}~(b), the higher value the dot-product is, which means the corresponding word is more likely to be a part of an aspect term. From this figure, we observe that the two real aspect terms hold much higher weights compared with the other words, which derives a correct detection. 
We can also obtain similar observations from the case shown as Figure~\ref{fig:aspectdetectioncase2} in which two sentiment capsules are active. In this case, there are two aspect terms, namely ``food'' and ``service''. 
From the sentence, we observe that this review has a strong negative sentiment towards on ``service'' while has a neutral attitude to ``food''. After we de-capsulize the active sentiment capsules, we find both ``service'' and ``food'' are perceived by the negative capsule because of the overall negative sentiment of the whole sentence. While in the neutral capsule, only the corresponding aspect term ``food'' is highlighted.


\section{Conclusion}\label{sec:conclusion}
In this paper, we proposed, CAPSAR, a capsule network based model for improving aspect-level sentiment analysis. 
The network is piled up hierarchical capsule layers equipped with a shared-weight routing algorithm to capture key features for predicting sentiment polarities. 
Meanwhile, the instantiation parameters of sentiment capsules are used to reconstruct the aspect representation, and the reconstruction loss is taken as a part of the training objective. As a consequence, CAPSAR could further capture the coherent patterns between sentiment and aspect information and is able to detect potential aspect terms by parsing the sentiment capsules when these aspect terms are unseen. 
Experimental results on three real-world benchmarks demonstrate the superiority of the proposed model.

\bibliographystyle{named}
\small{\bibliography{ijcai20}}

\end{document}